\documentclass[runningheads]{llncs}
\pdfoutput=1
 
\usepackage[mobile]{eccv}

\usepackage{eccvabbrv}

\usepackage{graphicx}
\usepackage{booktabs}

\usepackage[accsupp]{axessibility}  

\usepackage{hyperref}

\usepackage{orcidlink}

\begin{document}

\title{Gaussian in the Wild: 3D Gaussian Splatting for Unconstrained Image Collections} 

\titlerunning{Gaussian in the Wild}

\makeatletter
\newcommand{\printfnsymbol}[1]{%
  \textsuperscript{\@fnsymbol{#1}}%
}
\renewcommand*{\@fnsymbol}[1]{\ensuremath{\ifcase#1\or *\or \dagger\or \ddagger\or
   \mathsection\or \mathparagraph\or \|\or **\or \dagger\dagger
   \or \ddagger\ddagger \else\@ctrerr\fi}}
   
\makeatother
\author{Dongbin Zhang\thanks{Equal contributions.} \and
Chuming Wang\printfnsymbol{1} \and 
Weitao Wang\and
Peihao Li\and
Minghan Qin \and Haoqian Wang\thanks{Corresponding author.}}
\authorrunning{D.~Zhang, C.~Wang et al.}

\institute{Tsinghua Shenzhen International Graduate School, Tsinghua University}

\maketitle

\begin{figure}[h]
  \centering
  \includegraphics[width=\textwidth]{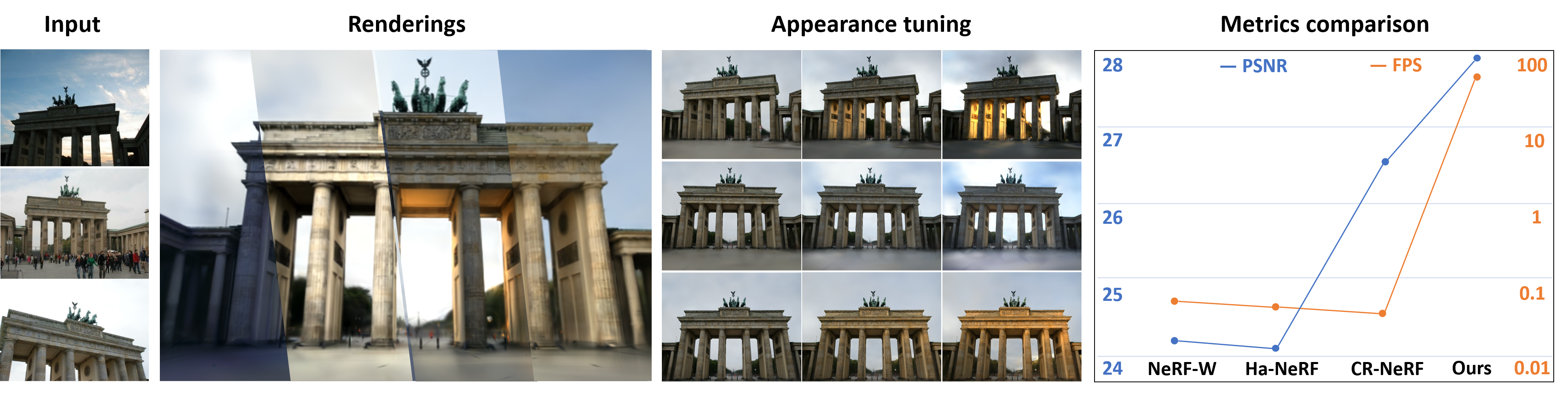}
 \caption{With an unconstrained image collection input, GS-W can render novel views with appearance tuning, achieving state-of-the-art quality and faster rendering speed.
  }
  \label{fig0}
\end{figure}
\begin{abstract}
Novel view synthesis from unconstrained in-the-wild images remains a meaningful but challenging task. The photometric variation and transient occluders in those unconstrained images make it difficult to reconstruct the original scene accurately. Previous approaches tackle the problem by introducing a global appearance feature in Neural Radiance Fields (NeRF). However, in the real world, the unique appearance of each tiny point in a scene is determined by its independent intrinsic material attributes and the varying environmental impacts it receives. Inspired by this fact, we propose Gaussian in the wild (GS-W), a method that uses 3D Gaussian points to reconstruct the scene and introduces separated intrinsic and dynamic appearance feature for each point, capturing the unchanged scene appearance along with dynamic variation like illumination and weather. Additionally, an adaptive sampling strategy is presented to allow each Gaussian point to focus on the local and detailed information more effectively. We also reduce the impact of transient occluders using a 2D visibility map. More experiments have demonstrated better reconstruction quality and details of GS-W compared to NeRF-based methods, with a faster rendering speed. Video results and code are available at \href{https://eastbeanzhang.github.io/GS-W/}{https://eastbeanzhang.github.io/GS-W/}.
\end{abstract}

\section{Introduction}
\label{sec:intro}
Novel view synthesis has long been a high-profile and complicated task in computer vision which aims to recover the 3D structure of a scene from 2D image collections and plays a significant role in many applications like virtual reality (VR) and autonomous driving. Recently, implicit representations, especially Neural Radiance Field (NeRF)\cite{mildenhall2021nerf} and its subsequent work \cite{barron2021mip,barron2022mip,muller2022instant} have shown impressive progress in rendering photorealistic images from arbitrary viewpoints. 

Meanwhile, explicit representations have also drawn increasing attention thanks to their real-time rendering speed. 3D Gaussian Splatting(3DGS)\cite{kerbl20233d} introduces 3D Gaussian as a novel and flexible scene representation and designs a fast differentiable rendering approach, allowing real-time rendering while maintaining high-fidelity for view synthesis.


However, these aforementioned methods only focus on static scenes from which images are captured without transient occluders and dynamic appearance variation such as ever-changing sky, weather, and illumination. Unfortunately, in practice, input unconstrained image collections may be captured by cameras with different settings at different times, and typically include pedestrians or vehicles. The assumption of the previous methods that images must be captured in static scenes is severely violated, resulting in sharp performance degradation\cite{martin2021nerf,yang2023cross}.

Several recent attempts \cite{martin2021nerf,chen2022hallucinated,tancik2022block,li2023nerf} adopt a global latent embedding for each image, granting their model the ability to handle the appearance variations between images. CR-NeRF\cite{yang2023cross} proposes a new cross-ray paradigm that utilizes information from multiple rays to obtain colors of several pixels, achieving more realistic and efficient appearance modeling. Nonetheless, the methods above still suffer from three defects:  \textbf{1) Global appearance representation:} The representation they use to control the dynamic appearance variation is shared by the whole scene, which may struggle to describe local high-frequency changes. By contrast, in the 3D real world, every single point of the scene has its appearance feature which brings them unique gloss and texture in different views;
\textbf{2) Blurring intrinsic and dynamic appearance:} An object’s intrinsic appearance is determined by its own material and surface properties while dynamic appearance is affected by environmental factors like highlight and shadow. Previous approaches blur the two appearances together, causing confusion in applications such as appearance tuning;
\textbf{3) High time cost:} Similar to volume rendering based methods, most NeRF-based approaches above suffer from high training costs and low rendering speed due to a large amount of network evaluation.

To address these challenges, we propose Gaussian in the wild (GS-W), a method to achieve high-quality and flexible scene reconstruction for unconstrained image collections. Specifically, we first use 3D Gaussian points to represent the scene and introduce independent appearance features to each point, enabling their unique appearance expression. Second, the intrinsic and dynamic appearance is separated and an adaptive sampling strategy is presented to grant every point the freedom to focus on various detailed dynamic appearance information. Additionally, the scene rendering speed is significantly accelerated thanks to the tile-based rasterizer.

Our contribution can be summarized as follows:
\begin{itemize}
    \item We propose a new framework GS-W, a 3D Gaussian Splatting based method, in which each Gaussian point is equipped with separated intrinsic and dynamic appearance features to enable more flexible varying appearance modeling from unconstrained image collections. 
    \item To better incorporate environmental factors from the image into the scene, we propose adaptive sampling, allowing each point to sample dynamic appearance features more effectively from the feature maps, thereby focusing on more local and detailed information. 
    \item Experimental results demonstrate that our method not only outperforms the state-of-the-art NeRF-based methods in terms of quality but also surpasses them in rendering speed by over $1000\times$.
\end{itemize}

\section{Related Work}
\subsection{3D representations}
Diverse 3D representations are developed to represent the geometric and appearance information of three-dimensional objects or scenes, among which implicit and explicit representation are two common methods for practical applications like 3D object generation and scene reconstruction. Implicit representation represents 3D data as continuous functions or fields, like occupancy fields\cite{mescheder2019occupancy}, distance fields \cite{park2019deepsdf}, color, and density. NeRF \cite{mildenhall2021nerf} is an outstanding work among them which models the scene as a continuous field of density and radiance. NeRF indicates that using MLP can represent complex scenes and render novel photo-realistic views with the help of volume rendering. On the contrary, explicit representation describes scenes by storing and manipulating 3D data using discrete structures like meshes\cite{wen2019pixel2mesh++,kanazawa2018learning,kato2018neural}, point clouds\cite{qi2017pointnet,qi2017pointnet++,shi2020pv}, voxels\cite{wu20153d,xu20223d,schwarz2022voxgraf} and so on. More recently, methods represented by 3DGS\cite{kerbl20233d} have entered researchers’ vision by their real-time rendering speed while preserving high-resolution synthesis quality. An efficient tile-based rasterizer is introduced by 3DGS to splat Gaussians to the image plane and accelerate the rendering speed many times compared to previous methods. Many researchers are exploring its potential by extending it to various tasks\cite{wu20234d,qin2023langsplat,yi2023gaussiandreamer}.  Several hybrid representations \cite{chan2022efficient,cao2023hexplane,chen2022tensorf,fridovich2023k,shao2023tensor4d} are also emerging and creating more possibilities.


\subsection{Novel view synthesis}
Synthesizing arbitrary views of a scene using a set of 2D images is a long-standing problem in computer vision. Many NeRF-based methods\cite{muller2022instant,fridovich2022plenoxels,yu2021plenoctrees,garbin2021fastnerf, reiser2021kilonerf} have achieved expressive synthesis quality \cite{barron2021mip,yang2023freenerf,verbin2022ref} along with good view-consistency\cite{deng2022depth, wang2023sparsenerf, niemeyer2022regnerf}. Mip-NeRF\cite{barron2021mip} replaces the rays with a 3D conical frustum and proposes integrated position embedding to anti-aliasing. Instant-NGP \cite{muller2022instant} introduces multi-resolution hash encoding that permits the use of a smaller network to reduce the training cost. There are also some Gaussian-based methods\cite{yang2024spec,lu2023scaffold,fan2023lightgaussian,yu2023mip} contributing to this task by modifying 3DGS. For example, to handle scenes with specular elements, Spec-Gaussian\cite{yang2024spec} departs from using spherical harmonics and instead adopts an anisotropic spherical Gaussian appearance field to model each point. Since these aforementioned methods all assume that input images are captured in a static scene, their performance declines intensely when reconstructing from unconstrained photo collections. Thus, several attempts\cite{meshry2019neural,martin2021nerf,rudnev2022nerf,chen2022hallucinated,yang2023cross,li2023nerf} are proposed to address this challenging in-the-wild task by handling appearance variation and transient occluders. Other works \cite{li2020crowdsampling,lin2023neural} focus on scenes with time-varying appearances, while methods \cite{zhang2021ners,kuang2022neroic,engelhardt2024shinobi} use physical rendering models for diverse lighting conditions. This field still faces some remaining issues and looks forward to advancements. 

As one of them, our proposed method tries to push the field one step forward by achieving a more delicate and flexible synthesis with higher rendering speed, through our modifications mentioned in \cref{4}.

\section{Preliminaries}
3D Gaussian Splatting (3DGS)\cite{kerbl20233d} is a method for reconstructing 3D scenes from static images with camera pose information. It uses explicit 3D Gaussian points $GP$ to represent the scene and achieves real-time image rendering through a differentiable tile-based rasterizer. These Gaussian points' positions $X$ are initialized with point clouds extracted by SFM\cite{schonberger2016structure} from the image set. Particularly, it uses 3D covariance $\Sigma$ to model the impact of each Gaussian point on the color anisotropy of the surrounding area:
\begin{equation}
    G(x-X,\Sigma)=e^{-\frac{1}{2}(x-X)^T\Sigma^{-1}(x-X) }
    \label{eq1}
\end{equation}

For ease of optimizing the covariance $\Sigma$ while maintaining its positive semi-definiteness, the method decomposes the covariance of each Gaussian point into a scaling matrix $S$ and a rotation matrix $R$, which are then stored as the Gaussian point attributes $s$ and $r$ respectively, using 3D vectors and quaternions.
\begin{equation}
    \Sigma=RSS^TR^T
    \label{eq2}
\end{equation}

Additionally, each Gaussian point is equipped with two more attributes: opacity $\alpha$ and color $c$, with the color attribute represented by third-order spherical harmonic coefficients. When rendering, besides projecting each Gaussian point onto a grid of $ 16\times16 $ tiles on the image plane, the 3D covariance $\Sigma$ is projected to 2D $\Sigma'$ using the viewing transformation $W$ and the Jacobian of the affine approximation of the projective transformation $J$:
\begin{equation}
    \Sigma'=JW\Sigma W^TJ^T
    \label{eq3}
\end{equation}

Then, based on the Gaussian points sorted by the rasterizer, the color of each pixel is aggregated using $\alpha$-blending:
\begin{align}
    \sigma_i&=G(px'-X_i,\Sigma'_i)   \label{eq4}\\
     C(px')&=\sum_{i\in GP_{px'}}c_i\sigma_i\prod_{j=1}^{i-1}(1-\sigma_j) \label{eq5}
\end{align}

Where $px'$ represents the position of a pixel, and $GP_{px'}$ denotes the sorted  Gaussian points associated with that pixel. The final rendered image is then used to compute loss with reference images for training, jointly optimizing all Gaussian attributes. Moreover, it devises a strategy for point growth and pruning based on gradients and opacity.

\begin{figure}[tb]
  \centering
  \includegraphics[width=\textwidth]{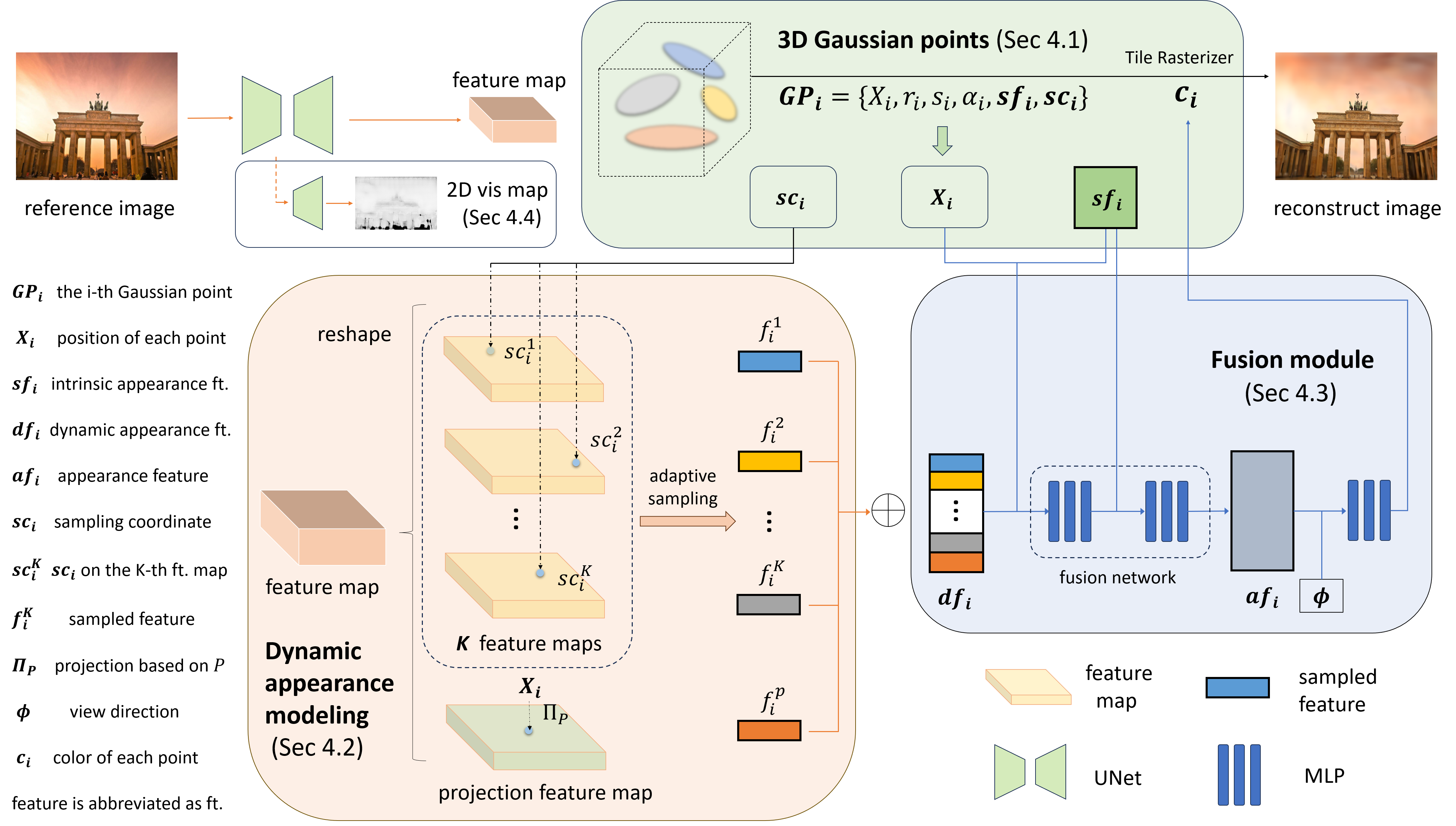}
  \caption{An overview of the GS-W framework. We begin with a scene's reference image and its camera pose $P$. After extracting image features via a Unet model, we reshape them into K feature maps and one projection feature map. Each Gaussian point $GP_i$ then samples features from these maps adaptively, capturing dynamic appearance feature $df_i$. These features are fused with the intrinsic appearance feature $sf_i$ through a fusion network, decoded for Gaussian point color $c_i$. Finally, all Gaussian points are rendered using a tile rasterizer. 
  }
  \label{fig1}
\end{figure}

\section{Method}
\label{4}
Based on previous analysis, aforementioned NeRF-based methods\cite{martin2021nerf,chen2022hallucinated,yang2023cross} lack enough attention to high-frequency and local detailed information in appearance, along with significant rendering costs due to the large number of sampling points. To address these issues, we utilize 3D Gaussian points to explicitly model the scene in \cref{4.1} and introduce a new appearance modeling method for each Gaussian point in \cref{4.2}. Intrinsic and dynamic appearance features are separated and then fused in \cref{4.3}. Additionally, when calculating the losses in \cref{4.5}, a visibility map is employed to reduce the impact of transient objects in \cref{4.4}. The whole pipeline is visualized as \cref{fig1}.
\subsection{3D Gaussian Splatting in the Wild}
\label{4.1}
\textbf{Appearance.} 
Since 3DGS\cite{kerbl20233d} is designed for reconstructing static scenes, GS-W abandons the conventional color modeling approach using spherical harmonic coefficients. Instead, in the subsequent section, we introduce a new appearance feature $af_i$ for each Gaussian point, adapting to variations in the reference image by fusing intrinsic appearance feature $sf_i$ with dynamic appearance feature $df_i$ extracted from the image.

\textbf{Transient object.} Handling transient objects is also challenging for 3DGS, in which Gaussian points around transient object regions may receive gradients and move or grow, bringing the emergence of meaningless floating points and rendering artifacts. Therefore, a visibility map is employed to mitigate this issue in \cref{4.4}. In addition, we maintain the pruning and growing strategies for points to reconstruct buildings that are absent in the initial point cloud.

\subsection{Dynamic appearance features modeling}
\label{4.2}

In the real physical world, even within the same scene, most object points experience varying environmental influences, such as light from different directions. Combined with unique intrinsic material properties, each point displays a different color, gloss, and texture. NeRF-W\cite{martin2021nerf} and Ha-NeRF\cite{chen2022hallucinated} employ a global feature embedding from the reference image for the entire scene's appearance, assuming uniform environmental information across all points. CR-NeRF\cite{yang2023cross} integrates reference image features into rendered image features at the 2D level, lacking dynamic environmental information in 3D. These methods tend to roughly restore the global color tone of the scene and struggle to capture local details such as highlight and shadow. To address this, we introduce diverse information for each point, better aligning with real-world scenarios.

\textbf{Projection feature map.} 
We extract 2D features from the image $I_{gt}$ and then map each point to the 2D space for feature sampling. With the known camera pose of the image, the mapping relationship from 3D points to the 2D image can be determined. Therefore, from the image we extract a projection feature map $F^P$, onto which each 3D point is projected using a projection matrix $P$, followed by the bilinear interpolation feature sampling, as illustrated below:
\begin{equation}
    \label{eq6}
    f_i^P=BL(\Pi_{P}(X_i),F^P)
\end{equation}

where $X_i$ represents the position coordinate of the i-th Gaussian point, while $f_i^P\in \mathbb{R}^{16}$ denotes the features sampled from the projection feature map for this Gaussian point, constituting a portion of its dynamic appearance feature. The term $BL$ indicates the bilinear interpolation sampling. 

By employing this method, Gaussian points along different rays can effectively capture features at their corresponding positions on the reference image.

\textbf{K feature maps.}
 Limited by the single-view reference image $I_{gt}$, features sampled from the projection feature map are identical along the same ray, which is unreasonable under uneven lighting. Besides, since the reference image sometimes only contains part of the scene, many points may not be projected into effective regions for obtaining valid feature samples, causing inconsistencies in new viewpoints. To address these issues and grant Gaussian points the freedom to focus on diverse information, we propose extracting additional $K$ feature maps $(F^1, F^2...F^K)$ from $I_{gt}$ to construct a high-dimensional sampling space. Each Gaussian point is mapped to these maps respectively for adaptive sampling, allowing them to focus on high-frequency features.

\textbf{Adaptive sampling.} It is necessary to map the Gaussian points to different positions on K feature maps, to better focus on diverse information. Inspired by the motive of each Gaussian point independently learning its own attributes, we consider it an efficient way to allow every point to determine its sampling positions through self-learning. Therefore, we assign each Gaussian point with K learnable sampling coordinate attributes $(sc^1_i,sc^2_i...sc^K_i)$, enabling them to adaptively select the information they need to focus on. The sampling process using these K sampling coordinates in K feature maps is as follows:
\begin{equation}
    \label{eq7}
    (f_i^1,f_i^2...f_i^K)=BL((sc_i^1,sc_i^2...sc_i^K),(F^1,F^2...F^K))
\end{equation}

where $(f_i^1,f_i^2...f_i^K)\in \mathbb{R}^{K\times16}$ represents the features sampled from K feature maps for the i-th Gaussian point. Next, we concatenate the sampled features with $f_i^P$ to jointly represent the dynamic appearance feature of the Gaussian point as follows:
\begin{equation}
    \label{eq8}
    df_i=f_i^P\oplus f_i^1\oplus f_i^2 \oplus... f_i^K
\end{equation}

During training, to prevent sampling coordinates from deviating beyond the effective sampling range, we introduce a regularization term:
\begin{equation}
    \label{eq9}
    L_{sc}=\frac{1}{N}\sum_{i}^{N} max\{0,|sc_i|-1\}
\end{equation}

where N represents the total number of Gaussian points, and $|.|$ denotes the absolute value.

\textbf{Feature maps extraction.} We utilize a Unet\cite{ronneberger2015u} model with ResNet\cite{he2016deep} backbone to generate both the projection feature map and K feature maps. The Unet model takes the reference image $I_{gt} \in \mathbb{R} ^{3\times H \times W}$ as input and produces a 2D feature map $F \in \mathbb{R}^{16(K+1) \times H \times W}$ of the same spatial size. Subsequently, this feature map is evenly divided along the channel dimension into $(K+1)$ feature maps, each serving as one of the K feature maps $(F^{1}, F^{2}...F^{K})$, and the projection feature map $F^{P}$. We choose Unet as the feature extractor due to its simplicity and effectiveness in extracting features from images at the same scale.

\subsection{Intrinsic and dynamic appearance}
\label{4.3}
\textbf{Separation of intrinsic and dynamic appearance.} 
As mentioned before, an object's appearance is influenced by both its intrinsic material as well as surface properties, and dynamic environmental factors. However, previous methods like Ha-NeRF primarily rely on image features, positional data, and viewing direction to decode appearance using MLPs. This implicit modeling of the scene's intrinsic attributes through MLPs makes it difficult to express the high-frequency features solely with small MLPs, thus relying on the extracted dynamic features to capture comprehensive information. Consequently, such blurring of features hinders the model's ability to distinguish between intrinsic and dynamic appearance accurately, particularly in scenarios involving changes in illumination and weather conditions. To address this, we explicitly separate the scene appearance into two forms: intrinsic and dynamic appearance features. 



Similar to modeling the static Gaussian points' positions, we assign a new learnable intrinsic appearance attribute $sf_i$ to each Gaussian point. Meanwhile, the dynamic appearance features $df_i$ are obtained by features extracted from the reference image.

\textbf{Fusion of intrinsic and dynamic appearance features.}
After acquiring independent dynamic appearance features for each Gaussian point through adaptive sampling, it's essential to combine them with the corresponding intrinsic appearance features to generate a comprehensive appearance feature $af_i$. To accomplish this, we design a fusion network $M_f$ composed of two MLPs, which takes two appearance features and position information as inputs and produces a holistic appearance feature influenced by both. Specifically:
\begin{equation}
    \label{eq10}
    af_i=M_f(sf_i,df_i,X_i)
\end{equation}

When rendering images, the fused appearance feature $af_i$, along with the view direction $\phi$, are jointly decoded by an MLP $M_c$ to obtain the color $c_i$ of the i-th Gaussian point, as in \cref{eq_11}. Finally, the Gaussian points are rendered using a differentiable tile rasterizer and colors are aggregated according to \cref{eq5}, producing image $I_r$ with the appearance features of the reference image.

\begin{equation}
    \label{eq_11}
 c_i=M_c(af_i,\phi)
\end{equation}

\subsection{Transient objects handling}
\label{4.4}
To mitigate the impact of transient objects and prevent the occurrence of artifacts like floating points, we employ a 2D visibility map $VM\in \mathbb{R}^{1\times H\times W}$ obtained from a Unet model, facilitating accurate segmentation between transient and static objects. Leveraging the visibility map, we weight the loss calculation between the reference image $I_{gt}$ and the rendered image $I_r$, as \cref{eq13}. During training, the model often struggles to reconstruct the geometry and appearance of transient objects, leading to larger losses in regions containing such objects. Therefore, in unsupervised scenarios, the 2D visibility map tends to decrease the visibility of transient objects to minimize the training loss. As a result, regions with higher visibility receive more emphasis, while those with lower visibility are disregarded. Additionally, to prevent the visibility map from marking all pixels invisible, we introduce a regularization loss term for the visibility map:
\begin{equation}
    \label{eq12}
    L_{vm}=L_2(VM,1)
\end{equation}

\subsection{Optimization}
\label{4.5}
Similar to \cite{kerbl20233d}, we apply two types of the loss function, $L_1$ and $L_{SSIM}$\cite{wang2004image}, to calculate the pixel error between the rendered image $I_r$ and the reference image $I_{gt}$. Differently, a visibility map $VM$ is incorporated to guide the supervision of rendered images by reference images. Moreover, we introduce the perceptual loss $L_{LPIPS}$\cite{zhang2018unreasonable}. Therefore, the overall image loss function is represented as in \cref{eq13}, where $\odot$ denotes the Hadamard product. Combining the regularization loss terms mentioned in \cref{eq9} and \cref{eq12}, the total loss function is formulated as in \cref{eq14}, where $\lambda_1$, $\lambda_{SSIM}$, $\lambda_{LPIPS}$, $\lambda_{sc}$ and $\lambda_{vm}$ are 0.8, 0.2, 0.005, 0.001, 0.15, respectively.

\begin{align}
    \label{eq13}
     L_c &= \lambda_{1}L_1(VM\odot I_r,VM\odot I_{gt})+\lambda_{SSIM}L_{SSIM}(VM\odot I_r,VM\odot I_{gt})  \nonumber  \\
     &+ \lambda_{LPIPS} L_{LPIPS}(I_r,I_{gt}) 
\end{align}

\begin{equation}
     \label{eq14}
      L=L_c+\lambda_{sc}L_{sc}+\lambda_{vm}L_{vm} 
\end{equation}

\section{Experiments}
\subsection{Implementation details} We implement our method using Pytorch \cite{paszke2019pytorch} and train our networks with Adam optimizer \cite{kingma2014adam}. We set $K=3$ in our experiment as the increasing K brings no performance improvement with meaningless computational cost. We train the full model on a single Nvidia RTX 3090 GPU for $70k$ steps and downsample all the images 2 times during training and evaluation, which takes approximately 2 hours. We also perform the adaptive control of the 3D Gaussians and follow other hyperparameter settings similar to 3DGS \cite{kerbl20233d}. 
\begin{table}[tb]
  \caption{Quantitative results on the test set of three PhotoTourism scenes. The \textbf{bold} and the \underline{underline} represent the best and second-best results, respectively. Our method outperforms the previous methods across all scenes on PSNR, SSIM, and LPIPS.  
  }
  \label{table1}
  \centering
  \begin{tabular}{@{}cccccccccc@{}}
     \toprule
     & \multicolumn{3}{c}{Brandenburg Gate} & \multicolumn{3}{c}{Sacre Coeur}  &  \multicolumn{3}{c}{Trevi Fountain} \\
      \cline{2-4}    \cline{5-7}     \cline{8-10} 
     & PSNR$\uparrow$ & SSIM$\uparrow$ & LPIPS$\downarrow$ & PSNR$\uparrow$ & SSIM$\uparrow$ & LPIPS$\downarrow$ & PSNR$\uparrow$ & SSIM$\uparrow$ & LPIPS$\downarrow$ \\
    \midrule
   3DGS   & 19.33 & 0.8838 & 0.1317 & 17.70 & \underline{0.8454} & 0.1761 & 17.08 & \underline{0.7139} & 0.2413 \\
   NeRF-W & 24.17 & 0.8905 & 0.1670 & 19.20 & 0.8076 & 0.1915 & 18.97 & 0.6984 & 0.2652 \\
   Ha-NeRF& 24.04 & 0.8873 & 0.1391 & 20.02 & 0.8012 & 0.1710 & 20.18 & 0.6908 & 0.2225 \\
   CR-NeRF& \underline{26.53} & \underline{0.9003} & \underline{0.1060} & \underline{22.07} & 0.8233 & \underline{0.1520} & \underline{21.48} & 0.7117 & \underline{0.2069} \\
   GS-W (Ours)   & \textbf{27.96} & \bf0.9319 & \bf0.0862 & \bf23.24 & \bf0.8632 & \bf0.1300 & \bf22.91 & \bf0.8014 & \bf0.1563 \\
   \bottomrule
  \end{tabular}
\end{table}

\begin{table}[tb]
  \caption{Comparison of rendering speed with one RTX 3090 GPU on three scenes with a resolution of 800 $\times$ 800, measured by FPS(Frames Per Second). Ours-cache means the appearance feature is cached for each point when synthesizing novel views.
  }
  \label{table2}
  \centering
  \begin{tabular}{@{}lccc@{}}
     \toprule
     & Brandenburg Gate &  \quad  \quad Sacre Coeur &  \quad \quad  Trevi Fountain \\
    \midrule
   3DGS   & 221 & 268 & 198 \\
   NeRF-W & 0.0518 & 0.0514 & 0.0485 \\
   Ha-NeRF& 0.0489 & 0.0497 & 0.0498 \\
   CR-NeRF& 0.0445 & 0.0447 & 0.0446  \\
   Ours &   55.8 &  58.3 &  38 \\
   Ours-cache  & 221 & 301  &  197 \\
   \bottomrule
  \end{tabular}
\end{table}

\subsection{Evaluation}
\textbf{Dataset, metrics, baseline.} We evaluate our proposed method on three scenes from the PhotoTourism dataset: Brandenburg Gate, Sacre Coeur, and Trevi Fountain, which include varying appearances and transient objects. For quantitative comparison, we use PSNR, SSIM\cite{wang2004image}, and LPIPS\cite{zhang2018unreasonable} as metrics to assess the performance of our method. We also present rendered images generated from the same pose as the input view for visual inspection. To demonstrate the superiority of our method, we evaluate our proposed method against 3DGS\cite{kerbl20233d}, NeRF-W \cite{martin2021nerf}, Ha-NeRF\cite{chen2022hallucinated}, and CR-NeRF\cite{yang2023cross}.

\textbf{Quantitative comparison.} Quantitative results are shown in \cref{table1}. 3DGS performs poorly on both PSNR and SSIM metrics, as it does not explicitly model the appearance variation and transient objects. NeRF-W and Ha-NeRF achieve moderate performance by introducing a global appearance embedding and anti-transient module. It's worth noting that NeRF-W needs to optimize the appearance embedding of test images, thus the comparison with NeRF-W is unfair. CR-NeRF achieves competitive performance due to cross-ray manner. Utilizing an adaptive sampling strategy that allows each point to focus on local details, our method outperforms the baselines on three scenes in terms of  PSNR, SSIM, and LPIPS, which verifies that we can capture more details and render higher-quality images.

\textbf{Render speed.} To compare the render speed of different methods during inference, we experiment on the three scenes by setting the image resolution to $800 \times 800$ and calculating the average rendering time per image using a single RTX 3090 GPU. The time taken for feature extraction from the reference images in Ha-NeRF, CR-NeRF, and our method is included in the overall inference time. As shown in \cref{table2}, our method achieves a significant improvement in rendering speed, which is 1000$\times$ faster than previous NeRF-based methods.

\begin{figure}[htbp]
  \centering
  \includegraphics[width=\textwidth]{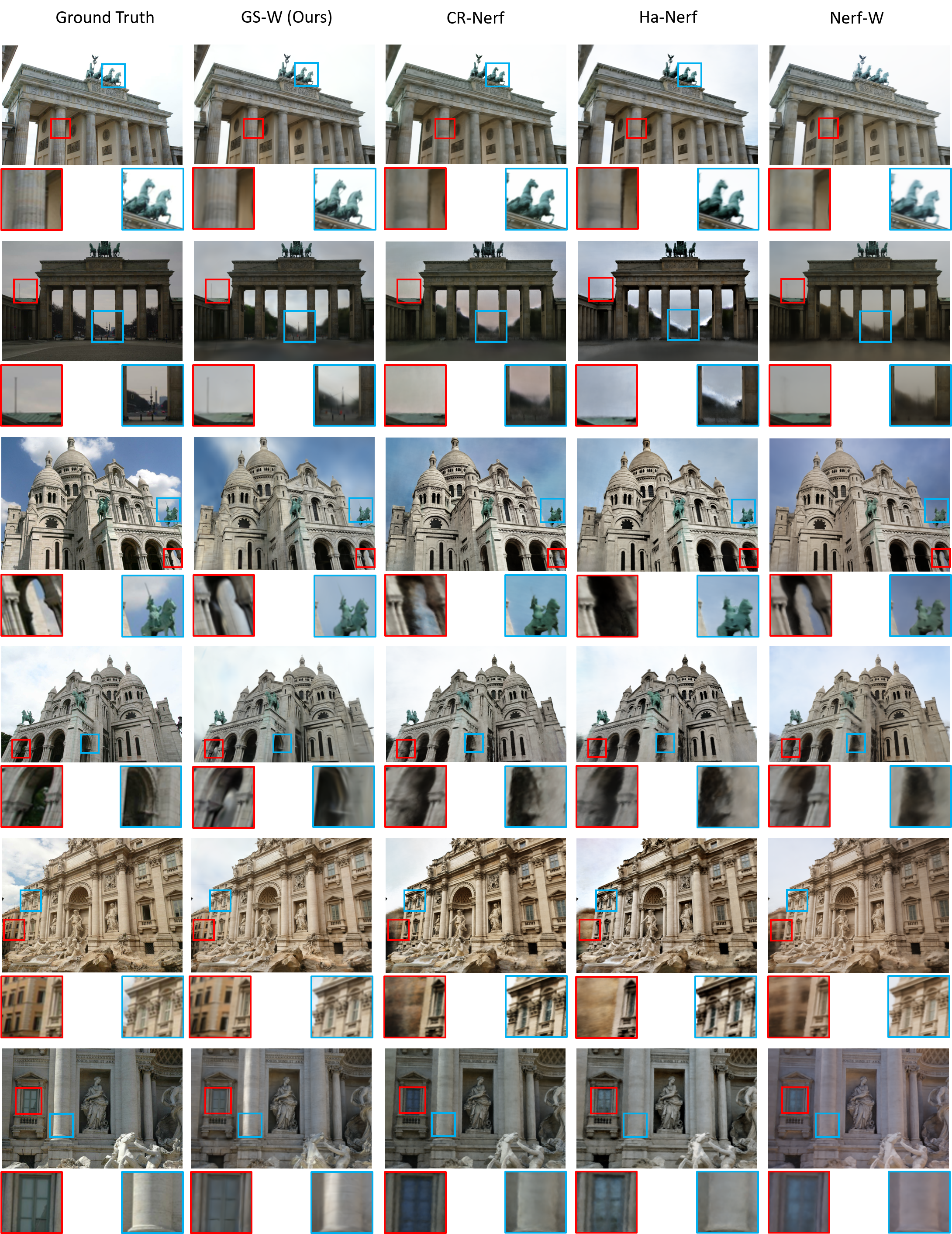}
  \caption{Qualitative results on the test set of three PhotoTourism scenes. GS-W recovers finer details of appearance(\eg{ the horse sculpture in Brandenburg, the sky and clouds in Sacre, the light on columns, and the color of windows in Trevi}). Moreover, GS-W reconstructs more consistent and detailed scenes (\eg{ the distant tower in Brandenburg, the cavities in Sacre, and the distant building in Trevi}).}
  \label{fig2}
\end{figure}

Since our method only requires one feature extraction step and can cache appearance feature $af_i$ for each Gaussian point, only a small MLP decoder $M_c$ is needed for color decoding when synthesizing a novel view. This enables GS-W to achieve a 200 FPS rendering speed, comparable to 3DGS.

\textbf{Qualitative comparison.} \cref{fig2} presents the qualitative result for all methods. NeRF-W and Ha-NeRF can model varying appearances from the reference image by introducing a global appearance embedding. CR-NeRF can reconstruct better geometry and model appearance variation compared with Ha-NeRF and NeRF-W. However, they all struggle to reconstruct the details of scenes in the distance and the intricate textures of the scenes, \eg{ the door pillars and distant tower in Brandenburg, the cavity in Sacre, and the distant buildings in Trevi}. In contrast, thanks to the higher-frequency dynamic appearance features, our method recovers more accurate appearance details, \eg{ the horse sculpture in Brandenburg, the sky and clouds in Sacre, and the light on columns in Trevi}.

\subsection{Ablation studies}
\label{5.3}
\begin{figure}[tb]
  \centering
  \includegraphics[width=\textwidth]{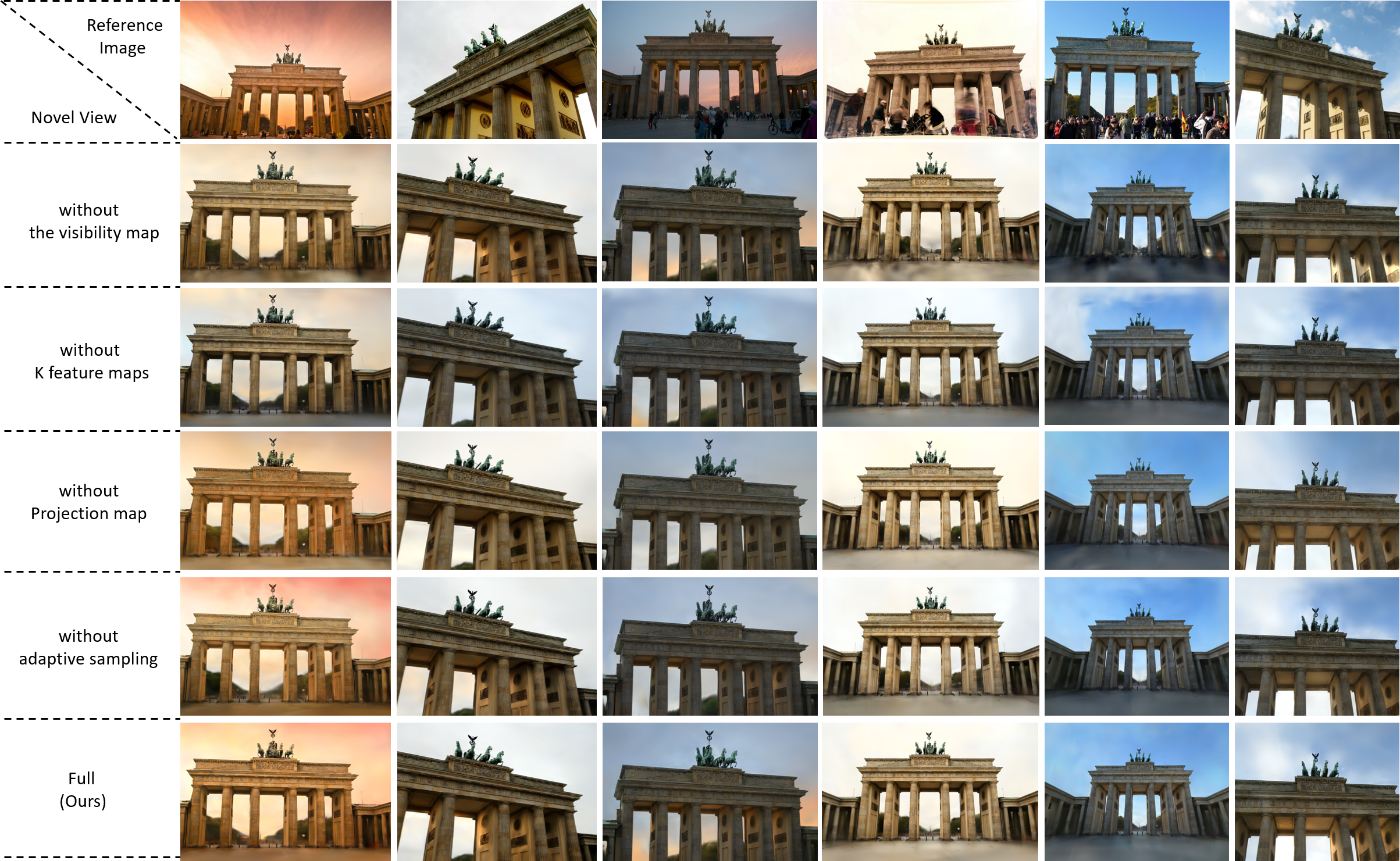}
  \caption{Ablation studies by visualization. The first row represents the reference images and the corresponding rows represent the rendered images from a novel view. Our full method is capable of performing view-consistent appearance and reducing artifacts.}
  \label{fig3}
\end{figure}

\begin{figure}[tb]
  \centering
  \includegraphics[width=\textwidth]{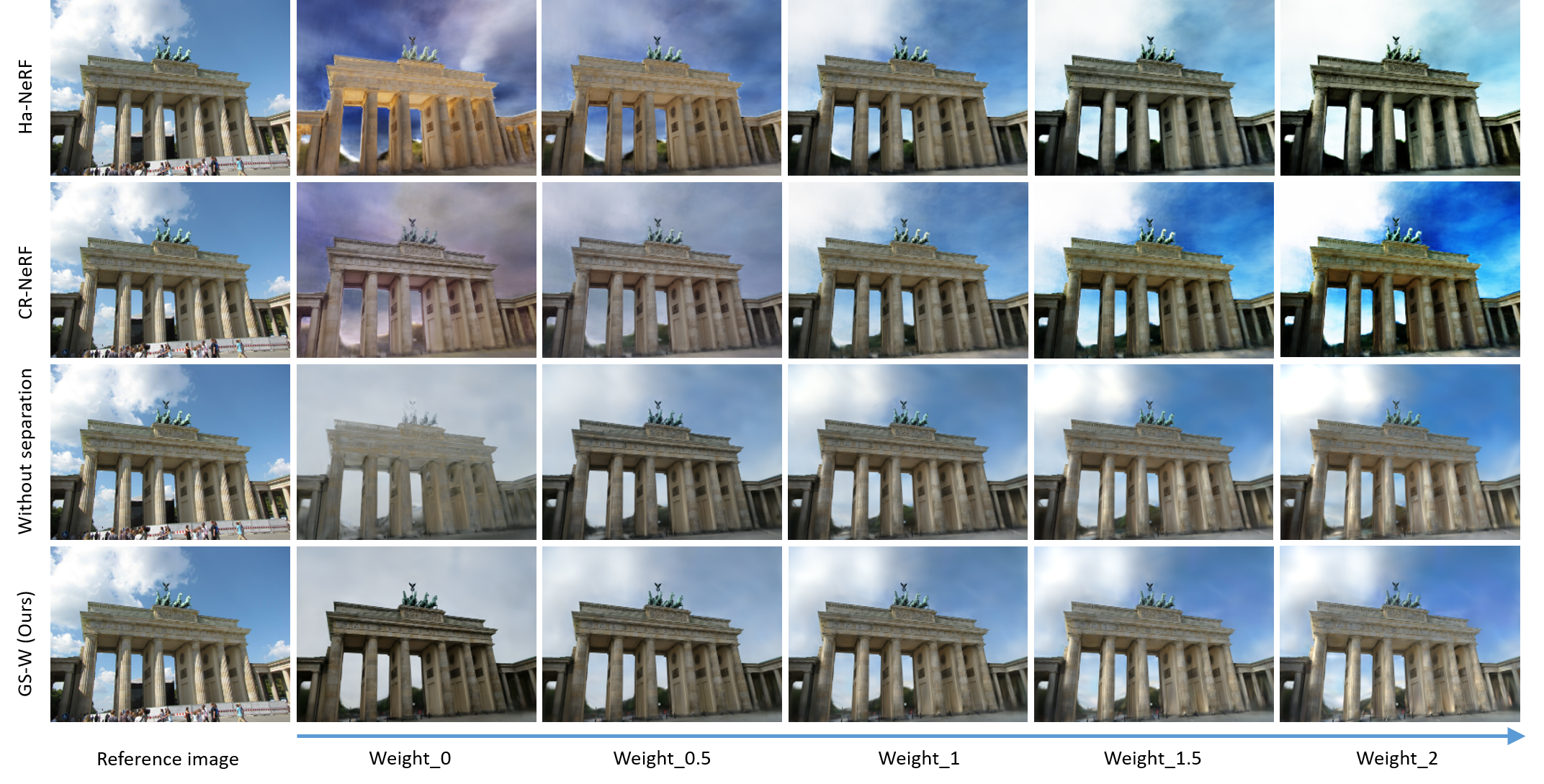}
  \caption{Images are rendered at the same camera pose with increasing weight of features extracted from the image. Our method incorporates environmental factors, like highlights on pillars and enhancing illumination, in a manner that is closer to human understanding.
  }
  \label{fig4}
\end{figure}

 We summarize the ablation studies of our method on Brandenburg, Sacre, and Trevi datasets in \cref{table3} and produce qualitative results in \cref{fig3} and \cref{fig4} to validate the effectiveness of each component. 
 
\textbf{Without the visibility map.} 
Removing the transient object handling module leads to higher metric performances but introduces artifacts in the rendered image due to the influence of transient objects, as depicted in \cref{fig3}. Since most test images lack dynamic objects, these artifacts may not significantly impact the metrics.


\textbf{Without K feature maps or projection feature map.} The removal of K feature maps or the projection feature map results in performance degradation. Especially, synthesizing novel views without the K feature maps not only reduces the ability to capture information from the reference image but also produces view-inconsistent appearances, as shown in \cref{fig3}.

\textbf{Without adaptive sampling.} 
Retaining K feature maps while keeping the sampling coordinates fixed for each point significantly decreases performance. This highlights the importance of our adaptive sampling strategy for K feature maps, enabling Gaussian points to adaptively focus on local and detailed features.

\begin{table}[tb]
  \caption{Ablation studies on three scenes. The \textbf{bold} and the \underline{underline} represent the best and second-best results, respectively. See \cref{5.3} for detailed descriptions.
  }
  \label{table3}
  \centering
  \resizebox{\textwidth}{!}{
  \begin{tabular}{@{}cccccccccc@{}}
     \toprule
     & \multicolumn{3}{c}{Brandenburg Gate} & \multicolumn{3}{c}{Sacre Coeur}  &  \multicolumn{3}{c}{Trevi Fountain} \\
      \cline{2-4}    \cline{5-7}     \cline{8-10} 
     & PSNR$\uparrow$ & SSIM$\uparrow$ & LPIPS$\downarrow$ & PSNR$\uparrow$ & SSIM$\uparrow$ & LPIPS$\downarrow$ & PSNR$\uparrow$ & SSIM$\uparrow$ & LPIPS$\downarrow$ \\
    \midrule
    w/o visibility map  & \textbf{28.64} & \textbf{0.9324} & \underline{0.0867} & \textbf{23.76} & \textbf{0.8723} & \textbf{0.1234} & \underline{23.03} & \underline{0.8025} & \textbf{0.1532} \\
   w/o K feature maps  & 26.37 & 0.9210 & 0.1018 & 21.59 & 0.8521 & 0.1417 & 22.27 & 0.7859 & 0.1700 \\
   w/o Projection map& 26.61 & 0.9285 & 0.0955 & 22.99 & 0.8635 & 0.1320 & 22.33 & 0.7980 & 0.1580 \\
   w/o adaptive sampling & 26.44 & 0.9219 & 0.0961 & 21.45 & 0.8479 & 0.1439 & 22.18 & 0.7866 & 0.1674 \\
   w/o separation & \underline{28.22} & 0.9291 & 0.0942 & 22.94 & 0.8520 & 0.1469 & \textbf{23.27} & 0.7907 & 0.1714 \\
  
   full (Ours) & 27.96 & \underline{0.9319} & \textbf{0.0862} & \underline{23.24} & \underline{0.8632} & \underline{0.1300} & 22.91 & \textbf{0.8014} & \underline{0.1563} \\
   \bottomrule
  \end{tabular}}
\end{table}

\textbf{Without separation.} We eliminate the intrinsic feature and predict the color of Gaussian points by the dynamic appearance feature only. The results in \cref{table3} show a noticeable decrease in both LPIPS and SSIM, which are more in line with human visual perception. In \cref{fig4}, the absence of dynamic appearance features results in incomplete scene appearance. This emphasizes the significance of intrinsic feature in retaining essential scene characteristics. Both illustrate that separation is beneficial for the model to accurately comprehend and reconstruct sharp appearances.

\subsection{Appearance tuning experiment}
\label{5.4}
Since we explicitly model the scene's appearance as unchanged intrinsic and varying dynamic features, we can adjust the impact of dynamic appearance features on intrinsic appearance by multiplying $df_i$ by a proportional weight. We also compare it with Ha-NeRF and CR-NeRF by applying the same weights to extracted image features. Qualitative results are shown in \cref{fig4}. Ha-NeRF and CR-NeRF show odd coloration in the sky and buildings at low weights, while at high weights, they fail to capture the detailed highlights and enhance the style of sufficient illumination, resulting in darker building colors. In contrast, our method aligns more closely with human understanding of the physical world. As the weight increases from small to large, our method gradually applies extracted environmental impacts to the scene, \eg{ appearing highlights on pillars and enhancing illumination}. This demonstrates the importance of dynamic features in capturing environmental information and explicitly separating intrinsic and dynamic appearances aid the model in clearly learning and distinguishing between the two, thereby achieving more flexible tuning over the scene's appearance.

\subsection{Limitations}
\label{5.5}
While GS-W outperforms previous methods, it still has limitations. It struggles with complex lighting variations, specular reflections, and accurately reconstructing textures in frequently occluded scenes, like the floor texture in the Brandenburg Gate scene. Additionally, it assumes known image poses when incorporating appearance information from reference images. Future research may focus on developing new appearance modeling techniques to address these issues.


\section{Conclusion}
In this paper, we introduce GS-W, a method for reconstructing scenes from unconstrained image collections. Using 3D Gaussian points as 3D representation, we introduce separated intrinsic and dynamic appearance features for each point to effectively model scene appearance. We propose an adaptive sampling strategy to capture local environmental factors like highlights and utilize a visibility map to handle transient objects. Our approach outperforms previous NeRF-based methods by providing better extraction of dynamic environmental impacts from images and addressing slow rendering speeds. Experimental results demonstrate the superiority and efficiency of our method compared to previous approaches.

\clearpage  

%
%
\bibliographystyle{splncs04}
\bibliography{main}

\begin{thebibliography}{10}
\providecommand{\url}[1]{\texttt{#1}}
\providecommand{\urlprefix}{URL }
\providecommand{\doi}[1]{https://doi.org/#1}

\bibitem{barron2021mip}
Barron, J.T., Mildenhall, B., Tancik, M., Hedman, P., Martin-Brualla, R., Srinivasan, P.P.: Mip-nerf: A multiscale representation for anti-aliasing neural radiance fields. In: Proceedings of the IEEE/CVF International Conference on Computer Vision. pp. 5855--5864 (2021)

\bibitem{barron2022mip}
Barron, J.T., Mildenhall, B., Verbin, D., Srinivasan, P.P., Hedman, P.: Mip-nerf 360: Unbounded anti-aliased neural radiance fields. In: Proceedings of the IEEE/CVF Conference on Computer Vision and Pattern Recognition. pp. 5470--5479 (2022)

\bibitem{cao2023hexplane}
Cao, A., Johnson, J.: Hexplane: A fast representation for dynamic scenes. In: Proceedings of the IEEE/CVF Conference on Computer Vision and Pattern Recognition. pp. 130--141 (2023)

\bibitem{chan2022efficient}
Chan, E.R., Lin, C.Z., Chan, M.A., Nagano, K., Pan, B., De~Mello, S., Gallo, O., Guibas, L.J., Tremblay, J., Khamis, S., et~al.: Efficient geometry-aware 3d generative adversarial networks. In: Proceedings of the IEEE/CVF Conference on Computer Vision and Pattern Recognition. pp. 16123--16133 (2022)

\bibitem{chen2022tensorf}
Chen, A., Xu, Z., Geiger, A., Yu, J., Su, H.: Tensorf: Tensorial radiance fields. In: European Conference on Computer Vision. pp. 333--350. Springer (2022)

\bibitem{chen2022hallucinated}
Chen, X., Zhang, Q., Li, X., Chen, Y., Feng, Y., Wang, X., Wang, J.: Hallucinated neural radiance fields in the wild. In: Proceedings of the IEEE/CVF Conference on Computer Vision and Pattern Recognition. pp. 12943--12952 (2022)

\bibitem{deng2022depth}
Deng, K., Liu, A., Zhu, J.Y., Ramanan, D.: Depth-supervised nerf: Fewer views and faster training for free. In: Proceedings of the IEEE/CVF Conference on Computer Vision and Pattern Recognition. pp. 12882--12891 (2022)

\bibitem{engelhardt2024shinobi}
Engelhardt, A., Raj, A., Boss, M., Zhang, Y., Kar, A., Li, Y., Sun, D., Brualla, R.M., Barron, J.T., Lensch, H., et~al.: Shinobi: Shape and illumination using neural object decomposition via brdf optimization in-the-wild. In: Proceedings of the IEEE/CVF Conference on Computer Vision and Pattern Recognition. pp. 19636--19646 (2024)

\bibitem{fan2023lightgaussian}
Fan, Z., Wang, K., Wen, K., Zhu, Z., Xu, D., Wang, Z.: Lightgaussian: Unbounded 3d gaussian compression with 15x reduction and 200+ fps. arXiv preprint arXiv:2311.17245  (2023)

\bibitem{fridovich2023k}
Fridovich-Keil, S., Meanti, G., Warburg, F.R., Recht, B., Kanazawa, A.: K-planes: Explicit radiance fields in space, time, and appearance. In: Proceedings of the IEEE/CVF Conference on Computer Vision and Pattern Recognition. pp. 12479--12488 (2023)

\bibitem{fridovich2022plenoxels}
Fridovich-Keil, S., Yu, A., Tancik, M., Chen, Q., Recht, B., Kanazawa, A.: Plenoxels: Radiance fields without neural networks. In: Proceedings of the IEEE/CVF Conference on Computer Vision and Pattern Recognition. pp. 5501--5510 (2022)

\bibitem{garbin2021fastnerf}
Garbin, S.J., Kowalski, M., Johnson, M., Shotton, J., Valentin, J.: Fastnerf: High-fidelity neural rendering at 200fps. In: Proceedings of the IEEE/CVF International Conference on Computer Vision. pp. 14346--14355 (2021)

\bibitem{he2016deep}
He, K., Zhang, X., Ren, S., Sun, J.: Deep residual learning for image recognition. In: Proceedings of the IEEE conference on computer vision and pattern recognition. pp. 770--778 (2016)

\bibitem{kanazawa2018learning}
Kanazawa, A., Tulsiani, S., Efros, A.A., Malik, J.: Learning category-specific mesh reconstruction from image collections. In: Proceedings of the European Conference on Computer Vision (ECCV). pp. 371--386 (2018)

\bibitem{kato2018neural}
Kato, H., Ushiku, Y., Harada, T.: Neural 3d mesh renderer. In: Proceedings of the IEEE conference on computer vision and pattern recognition. pp. 3907--3916 (2018)

\bibitem{kerbl20233d}
Kerbl, B., Kopanas, G., Leimk{\"u}hler, T., Drettakis, G.: 3d gaussian splatting for real-time radiance field rendering. ACM Transactions on Graphics  \textbf{42}(4) (2023)

\bibitem{kingma2014adam}
Kingma, D.P., Ba, J.: Adam: A method for stochastic optimization. arXiv preprint arXiv:1412.6980  (2014)

\bibitem{kuang2022neroic}
Kuang, Z., Olszewski, K., Chai, M., Huang, Z., Achlioptas, P., Tulyakov, S.: Neroic: Neural rendering of objects from online image collections. ACM Transactions on Graphics (TOG)  \textbf{41}(4),  1--12 (2022)

\bibitem{li2023nerf}
Li, P., Wang, S., Yang, C., Liu, B., Qiu, W., Wang, H.: Nerf-ms: Neural radiance fields with multi-sequence. In: Proceedings of the IEEE/CVF International Conference on Computer Vision. pp. 18591--18600 (2023)

\bibitem{li2020crowdsampling}
Li, Z., Xian, W., Davis, A., Snavely, N.: Crowdsampling the plenoptic function. In: Computer Vision--ECCV 2020: 16th European Conference, Glasgow, UK, August 23--28, 2020, Proceedings, Part I 16. pp. 178--196. Springer (2020)

\bibitem{lin2023neural}
Lin, H., Wang, Q., Cai, R., Peng, S., Averbuch-Elor, H., Zhou, X., Snavely, N.: Neural scene chronology. In: Proceedings of the IEEE/CVF Conference on Computer Vision and Pattern Recognition. pp. 20752--20761 (2023)

\bibitem{lu2023scaffold}
Lu, T., Yu, M., Xu, L., Xiangli, Y., Wang, L., Lin, D., Dai, B.: Scaffold-gs: Structured 3d gaussians for view-adaptive rendering. arXiv preprint arXiv:2312.00109  (2023)

\bibitem{martin2021nerf}
Martin-Brualla, R., Radwan, N., Sajjadi, M.S., Barron, J.T., Dosovitskiy, A., Duckworth, D.: Nerf in the wild: Neural radiance fields for unconstrained photo collections. In: Proceedings of the IEEE/CVF Conference on Computer Vision and Pattern Recognition. pp. 7210--7219 (2021)

\bibitem{mescheder2019occupancy}
Mescheder, L., Oechsle, M., Niemeyer, M., Nowozin, S., Geiger, A.: Occupancy networks: Learning 3d reconstruction in function space. In: Proceedings of the IEEE/CVF conference on computer vision and pattern recognition. pp. 4460--4470 (2019)

\bibitem{meshry2019neural}
Meshry, M., Goldman, D.B., Khamis, S., Hoppe, H., Pandey, R., Snavely, N., Martin-Brualla, R.: Neural rerendering in the wild. In: Proceedings of the IEEE/CVF Conference on Computer Vision and Pattern Recognition. pp. 6878--6887 (2019)

\bibitem{mildenhall2021nerf}
Mildenhall, B., Srinivasan, P.P., Tancik, M., Barron, J.T., Ramamoorthi, R., Ng, R.: Nerf: Representing scenes as neural radiance fields for view synthesis. Communications of the ACM  \textbf{65}(1),  99--106 (2021)

\bibitem{muller2022instant}
M{\"u}ller, T., Evans, A., Schied, C., Keller, A.: Instant neural graphics primitives with a multiresolution hash encoding. ACM Transactions on Graphics (ToG)  \textbf{41}(4),  1--15 (2022)

\bibitem{niemeyer2022regnerf}
Niemeyer, M., Barron, J.T., Mildenhall, B., Sajjadi, M.S., Geiger, A., Radwan, N.: Regnerf: Regularizing neural radiance fields for view synthesis from sparse inputs. In: Proceedings of the IEEE/CVF Conference on Computer Vision and Pattern Recognition. pp. 5480--5490 (2022)

\bibitem{park2019deepsdf}
Park, J.J., Florence, P., Straub, J., Newcombe, R., Lovegrove, S.: Deepsdf: Learning continuous signed distance functions for shape representation. In: Proceedings of the IEEE/CVF conference on computer vision and pattern recognition. pp. 165--174 (2019)

\bibitem{paszke2019pytorch}
Paszke, A., Gross, S., Massa, F., Lerer, A., Bradbury, J., Chanan, G., Killeen, T., Lin, Z., Gimelshein, N., Antiga, L., et~al.: Pytorch: An imperative style, high-performance deep learning library. Advances in neural information processing systems  \textbf{32} (2019)

\bibitem{qi2017pointnet}
Qi, C.R., Su, H., Mo, K., Guibas, L.J.: Pointnet: Deep learning on point sets for 3d classification and segmentation. In: Proceedings of the IEEE conference on computer vision and pattern recognition. pp. 652--660 (2017)

\bibitem{qi2017pointnet++}
Qi, C.R., Yi, L., Su, H., Guibas, L.J.: Pointnet++: Deep hierarchical feature learning on point sets in a metric space. Advances in neural information processing systems  \textbf{30} (2017)

\bibitem{qin2023langsplat}
Qin, M., Li, W., Zhou, J., Wang, H., Pfister, H.: Langsplat: 3d language gaussian splatting. arXiv preprint arXiv:2312.16084  (2023)

\bibitem{reiser2021kilonerf}
Reiser, C., Peng, S., Liao, Y., Geiger, A.: Kilonerf: Speeding up neural radiance fields with thousands of tiny mlps. In: Proceedings of the IEEE/CVF International Conference on Computer Vision. pp. 14335--14345 (2021)

\bibitem{ronneberger2015u}
Ronneberger, O., Fischer, P., Brox, T.: U-net: Convolutional networks for biomedical image segmentation. In: Medical Image Computing and Computer-Assisted Intervention--MICCAI 2015: 18th International Conference, Munich, Germany, October 5-9, 2015, Proceedings, Part III 18. pp. 234--241. Springer (2015)

\bibitem{rudnev2022nerf}
Rudnev, V., Elgharib, M., Smith, W., Liu, L., Golyanik, V., Theobalt, C.: Nerf for outdoor scene relighting. In: European Conference on Computer Vision. pp. 615--631. Springer (2022)

\bibitem{schonberger2016structure}
Schonberger, J.L., Frahm, J.M.: Structure-from-motion revisited. In: Proceedings of the IEEE conference on computer vision and pattern recognition. pp. 4104--4113 (2016)

\bibitem{schwarz2022voxgraf}
Schwarz, K., Sauer, A., Niemeyer, M., Liao, Y., Geiger, A.: Voxgraf: Fast 3d-aware image synthesis with sparse voxel grids. Advances in Neural Information Processing Systems  \textbf{35},  33999--34011 (2022)

\bibitem{shao2023tensor4d}
Shao, R., Zheng, Z., Tu, H., Liu, B., Zhang, H., Liu, Y.: Tensor4d: Efficient neural 4d decomposition for high-fidelity dynamic reconstruction and rendering. In: Proceedings of the IEEE/CVF Conference on Computer Vision and Pattern Recognition. pp. 16632--16642 (2023)

\bibitem{shi2020pv}
Shi, S., Guo, C., Jiang, L., Wang, Z., Shi, J., Wang, X., Li, H.: Pv-rcnn: Point-voxel feature set abstraction for 3d object detection. In: Proceedings of the IEEE/CVF conference on computer vision and pattern recognition. pp. 10529--10538 (2020)

\bibitem{tancik2022block}
Tancik, M., Casser, V., Yan, X., Pradhan, S., Mildenhall, B., Srinivasan, P.P., Barron, J.T., Kretzschmar, H.: Block-nerf: Scalable large scene neural view synthesis. In: Proceedings of the IEEE/CVF Conference on Computer Vision and Pattern Recognition. pp. 8248--8258 (2022)

\bibitem{verbin2022ref}
Verbin, D., Hedman, P., Mildenhall, B., Zickler, T., Barron, J.T., Srinivasan, P.P.: Ref-nerf: Structured view-dependent appearance for neural radiance fields. In: 2022 IEEE/CVF Conference on Computer Vision and Pattern Recognition (CVPR). pp. 5481--5490. IEEE (2022)

\bibitem{wang2023sparsenerf}
Wang, G., Chen, Z., Loy, C.C., Liu, Z.: Sparsenerf: Distilling depth ranking for few-shot novel view synthesis. arXiv preprint arXiv:2303.16196  (2023)

\bibitem{wang2004image}
Wang, Z., Bovik, A.C., Sheikh, H.R., Simoncelli, E.P.: Image quality assessment: from error visibility to structural similarity. IEEE transactions on image processing  \textbf{13}(4),  600--612 (2004)

\bibitem{wen2019pixel2mesh++}
Wen, C., Zhang, Y., Li, Z., Fu, Y.: Pixel2mesh++: Multi-view 3d mesh generation via deformation. In: Proceedings of the IEEE/CVF international conference on computer vision. pp. 1042--1051 (2019)

\bibitem{wu20234d}
Wu, G., Yi, T., Fang, J., Xie, L., Zhang, X., Wei, W., Liu, W., Tian, Q., Wang, X.: 4d gaussian splatting for real-time dynamic scene rendering. arXiv preprint arXiv:2310.08528  (2023)

\bibitem{wu20153d}
Wu, Z., Song, S., Khosla, A., Yu, F., Zhang, L., Tang, X., Xiao, J.: 3d shapenets: A deep representation for volumetric shapes. In: Proceedings of the IEEE conference on computer vision and pattern recognition. pp. 1912--1920 (2015)

\bibitem{xu20223d}
Xu, Y., Peng, S., Yang, C., Shen, Y., Zhou, B.: 3d-aware image synthesis via learning structural and textural representations. In: Proceedings of the IEEE/CVF Conference on Computer Vision and Pattern Recognition. pp. 18430--18439 (2022)

\bibitem{yang2023freenerf}
Yang, J., Pavone, M., Wang, Y.: Freenerf: Improving few-shot neural rendering with free frequency regularization. In: Proceedings of the IEEE/CVF Conference on Computer Vision and Pattern Recognition. pp. 8254--8263 (2023)

\bibitem{yang2023cross}
Yang, Y., Zhang, S., Huang, Z., Zhang, Y., Tan, M.: Cross-ray neural radiance fields for novel-view synthesis from unconstrained image collections. In: Proceedings of the IEEE/CVF International Conference on Computer Vision. pp. 15901--15911 (2023)

\bibitem{yang2024spec}
Yang, Z., Gao, X., Sun, Y., Huang, Y., Lyu, X., Zhou, W., Jiao, S., Qi, X., Jin, X.: Spec-gaussian: Anisotropic view-dependent appearance for 3d gaussian splatting. arXiv preprint arXiv:2402.15870  (2024)

\bibitem{yi2023gaussiandreamer}
Yi, T., Fang, J., Wu, G., Xie, L., Zhang, X., Liu, W., Tian, Q., Wang, X.: Gaussiandreamer: Fast generation from text to 3d gaussian splatting with point cloud priors. arXiv preprint arXiv:2310.08529  (2023)

\bibitem{yu2021plenoctrees}
Yu, A., Li, R., Tancik, M., Li, H., Ng, R., Kanazawa, A.: Plenoctrees for real-time rendering of neural radiance fields. In: Proceedings of the IEEE/CVF International Conference on Computer Vision. pp. 5752--5761 (2021)

\bibitem{yu2023mip}
Yu, Z., Chen, A., Huang, B., Sattler, T., Geiger, A.: Mip-splatting: Alias-free 3d gaussian splatting. arXiv preprint arXiv:2311.16493  (2023)

\bibitem{zhang2021ners}
Zhang, J., Yang, G., Tulsiani, S., Ramanan, D.: Ners: Neural reflectance surfaces for sparse-view 3d reconstruction in the wild. Advances in Neural Information Processing Systems  \textbf{34},  29835--29847 (2021)

\bibitem{zhang2018unreasonable}
Zhang, R., Isola, P., Efros, A.A., Shechtman, E., Wang, O.: The unreasonable effectiveness of deep features as a perceptual metric. In: Proceedings of the IEEE conference on computer vision and pattern recognition. pp. 586--595 (2018)

\end{thebibliography}

\newpage
\title{Supplementary material for \textquotedblleft Gaussian in the Wild\textquotedblright}
\author{}
\institute{}
\titlerunning{Gaussian in the Wild}
\maketitle
\setcounter{section}{0}
\renewcommand\thesection{\Alph  {section}}

\section {Video demo}
We strongly recommend readers to watch the video demo provided in \href{https://eastbeanzhang.github.io/GS-W/}{\textcolor{cyan}{webpage}}. The video showcases the novel view synthesis achieved by GS-W across various scenes, based on different reference images. It also illustrates the variations in scene appearance attained through the interpolation of different dynamic features $df$. Additionally, we provide a visual comparison between GS-W and CR-NeRF in terms of novel view synthesis and appearance tuning. The video illustrates GS-W's superior reconstruction of scene geometry and precise capture of environmental factors from the reference images, maintaining multi-view consistency. Furthermore, it highlights GS-W's ability to adjust scene appearance in a manner more aligned with human perception by weighting the features extracted from the reference images.
\section{More implementation details}
\subsection {Model parameters}
Below, we provide a detailed overview of the network parameters and other hyperparameter settings in GS-W. The Unet's encoder utilizes the first 8 sub-convolution modules from a pre-trained ResNet-18 to extract image features. On the other hand, the decoder for generating $K+1$ feature maps comprises 4 up-sampling convolution modules, along with a decoder including 3 up-sampling convolution modules for producing the visibility map. The two decoder modules are both equipped with batch normalization and ReLU activation functions. To reduce computation, there are no skip connections between the encoder and the visibility map generating decoder during the forward process. Transpose convolution is employed for up-sampling, with the number of channels for each feature map fixed at 16. The learning rate for the Unet network gradually decreases from  $ 2 \times 10^{-3}$ to $2 \times 10^{-5}$.

The Fusion network  $M_f$ consists of two MLP modules, each with 3 and 2 hidden layers, and each with hidden unit counts of $[128, 96, 64]$ and $[48, 48]$,  respectively. Meanwhile, the MLP $M_c$ for color decoding comprises only one hidden layer with 48 units. Xavier initialization is used for initializing the weight parameters of these MLPs. During the forward process, we use 10 frequencies to encode position. For training, random dropout with 0.1 probability is applied to the dynamic appearance features $df_i$ to prevent overfitting, and the learning rate for training these MLPs is set to $5 \times 10^{-4}$. The dimension of the intrinsic feature $sf_i$ is set to 48. 

Furthermore, the learning rate for Gaussian point positions decreases from  $1.6 \times 10^{-4}$ to $1.6 \times 10^{-7}$. Gaussian points are densified from 500 iterations to 15k iterations during training, with a gradient threshold set to $4 \times 10^{-4}$. Other hyperparameters are set according to the guidelines of 3DGS.

\subsection {Initialization of sampling coordinates}
In \cref{4.2}, we equip each Gaussian point with K learnable sampling coordinate attributes $(sc^1_i, sc^2_i, ..., sc^K_i) \in \mathbb{R}^{K \times 2}$ to enable it to adaptively sample features on K feature maps. Specifically, here we describe how to initialize them before training. To ensure the points close to each other have similar initial sampling coordinates, we utilize linear transformation to convert the position $X_i \in \mathbb{R}^3$ of each Gaussian point into K sampling coordinates. First, we randomly generate K matrices $(M^1, M^2, ..., M^K) \in \mathbb{R}^{K \times 2 \times 3}$,  while ensuring that the sum of each matrix's rows is 1. Then, we multiply these matrices with each Gaussian point's position to obtain the initial values of these K sampling coordinates, as follows: 
\begin{equation}
     \label{eq15}
      (sc^1_i, sc^2_i, ..., sc^K_i)_{init} = (M^1X_i, M^2X_i, ..., M^KX_i)
\end{equation}

\begin{table}[tb]
    \caption{Average quantitative results on the Brandenburg, Sacre, and Trevi scenes.}
    \label{table_s3}
  \centering
    \resizebox{0.6\linewidth}{!}{
  \begin{tabular}{@{}cccccc@{}}
     \toprule
     & PSNR$\uparrow$ \quad  & SSIM$\uparrow$ \quad  & LPIPS$\downarrow$ \quad  & FPS$\uparrow$ \quad  & Size(MB) \\
    \midrule
    global feature  \quad  & 22.43      \quad        & 0.8555      \quad & 0.1349 \quad &\textbf{56.4} \quad & 123.0\\
    K-planes-hybrid \quad   & 22.40 \quad  & 0.7604  \quad        & 0.2731 \quad &0.67 \quad &133.3\\
    Ours \quad  & \textbf{24.70} \quad    & \textbf{0.8655}   \quad   & \textbf{0.1242} \quad &50.7 \quad &122.7\\
    
   \bottomrule
  \end{tabular}}
    
\end{table}

\begin{figure}[t]
  \centering
  \includegraphics[width=\textwidth]{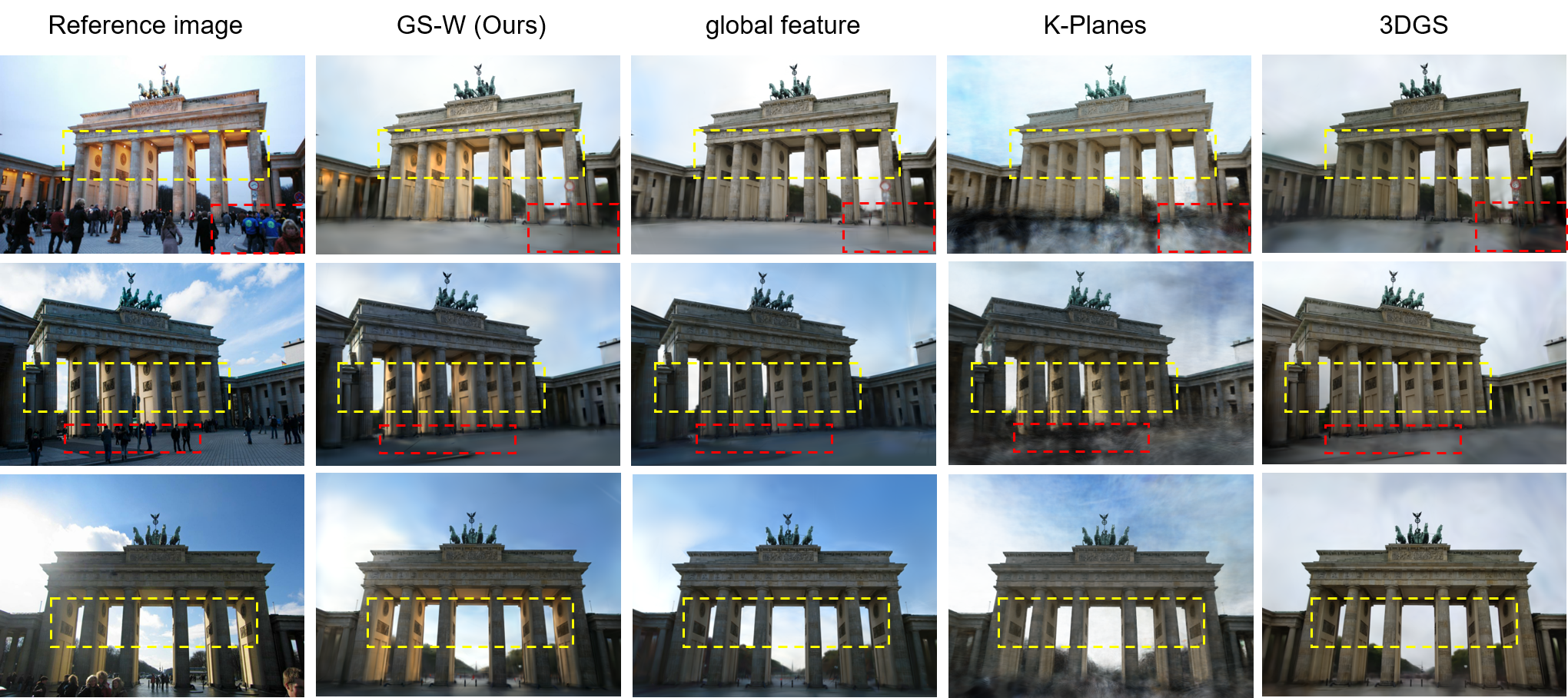}
  \caption{More qualitative comparison results on Brandenburg. Our method captures local highlights and avoids artifacts caused by transient objects, as observed in K-Planes and 3DGS.
  }
  \label{fig_s4}
\end{figure}

\begin{table}[tb]
  \caption{Quantitative results on the Sacre Coeur for different K values, where K represents the number of feature maps. We Choose K = 3 to achieve efficiency and effectiveness.}
  \label{table_s1}
  \centering
    \resizebox{0.6\textwidth}{!}{
  \begin{tabular}{@{}cccccc@{}}
     \toprule
      & \multicolumn{5}{c}{Sacre Coeur}  \\
      \cline{2-6}    
     & PSNR$\uparrow$ \quad \quad & SSIM$\uparrow$ \quad \quad & LPIPS$\downarrow$ \quad \quad & FPS$\uparrow$ \quad \quad &Size(MB)\\
    \midrule
    K=1 \quad \quad  & 23.08     \quad \quad        & 0.8598     \quad \quad       & 0.1318 \quad \quad & \textbf{60.3} \quad \quad & 93.8  \\
    K=2 \quad \quad  & 23.13 \quad \quad & 0.8590    \quad \quad & 0.1334 \quad \quad & 59.6 \quad \quad & 96.1\\
    K=3 \quad \quad & \textbf{23.24} \quad \quad   & \textbf{0.8632}  \quad \quad   & 0.1300 \quad \quad & 58.3 \quad \quad & 97.2\\
    K=4 \quad \quad & 23.02    \quad \quad        &0.8616 \quad \quad & 0.1303 \quad \quad & 58.0 \quad \quad & 98.2\\
    K=8 \quad \quad & 22.94    \quad \quad        &0.8618 \quad \quad & \textbf{0.1296} \quad \quad & 55.8 \quad \quad & 101.8\\
    K=16 \quad \quad & 22.98    \quad \quad        &0.8611 \quad \quad & 0.1321 \quad \quad & 51.8 \quad \quad & 113.9\\
   \bottomrule
  \end{tabular}}
\end{table}

\begin{figure}[tb]
  \centering
  \includegraphics[width=\textwidth]{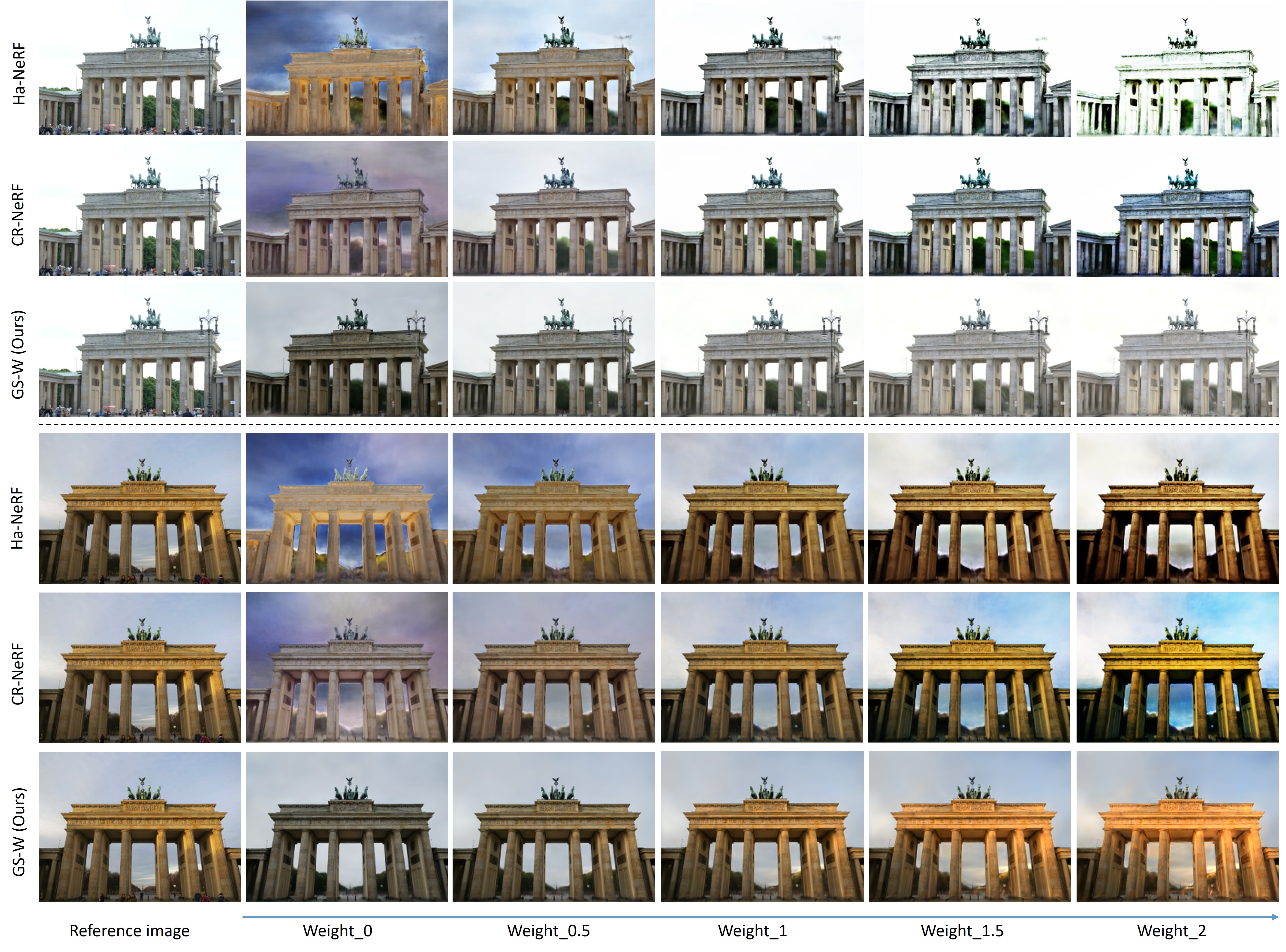}
  \caption{More qualitative comparison experiments on appearance tuning. Similar to \cref{fig4}, images are rendered at the same camera pose with increasing weight of features extracted from the image. Our method not only captures environmental information better but also naturally adjusts its influence on the scene.
  }
  \label{fig_s1}
\end{figure}

\begin{table}[t]
  \caption{Quantitative results on the synthetic dataset with introduced perturbations (colors \& occluders). GS-W outperforms the others in all evaluation metrics.
  }
  \label{table_s2}
  \centering
    \resizebox{0.5\textwidth}{!}{
  \begin{tabular}{@{}cccc@{}}
     \toprule
      & \multicolumn{3}{c}{Synthetic Lego Dataset}  \\
      \cline{2-4}    
     & PSNR$\uparrow$ \quad \quad & SSIM$\uparrow$ \quad \quad & LPIPS$\downarrow$ \\
    \midrule
    3DGS \quad \quad & 23.73  \quad \quad  & 0.9250   \quad \quad & 0.0748\\
    NeRF-W \quad \quad & 26.76 \quad \quad & 0.9224       \quad \quad     & 0.0552\\
    Ha-NeRF \quad \quad  & 26.51  \quad \quad  & 0.9416  \quad \quad  & 0.0339\\
    GS-W (Ours) \quad \quad & \textbf{29.64}   \quad \quad         & \textbf{0.9522} \quad \quad & \textbf{0.0323}\\
   \bottomrule
  \end{tabular}}
\end{table}

\begin{figure}[tb]
  \centering
  \includegraphics[width=\textwidth]{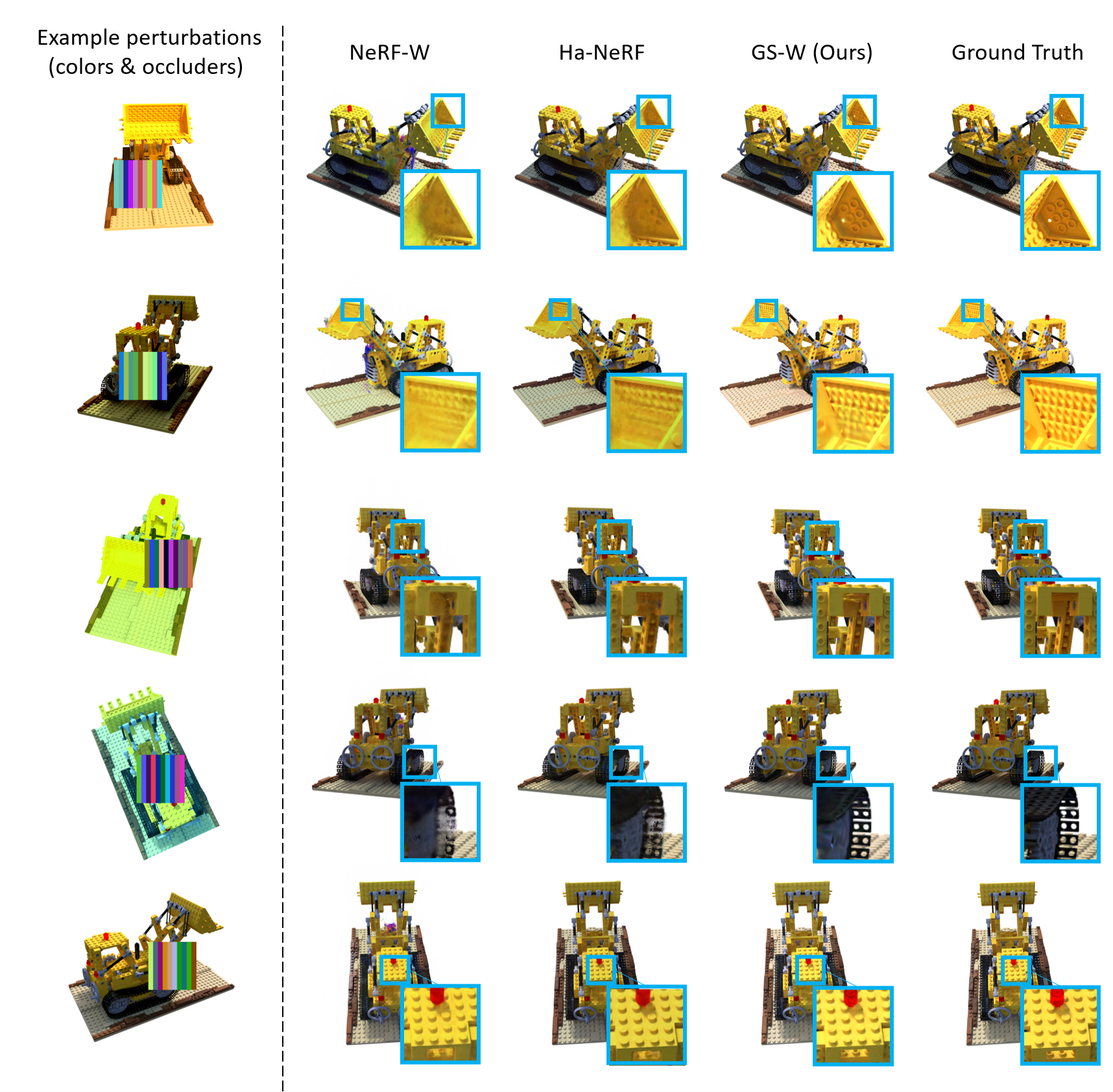}
  \caption{
Experiment results on synthetic Lego dataset with perturbations (colors \& occluders). Our method excels in modeling finer details and achieving more accurate color.
  }
  \label{fig_s2}
\end{figure}

\section{Further experiments}

\subsection {More comparison}
\textbf{Global appearance feature.} To highlight the superiority of independently sampled dynamic appearance features over a single global feature, We replace each point's dynamic appearance feature with a global one extracted from the reference image. As shown in \cref{table_s3} and \cref{fig_s4}, the global feature method achieves an inferior result and fails to recover local appearance details on the pillars and sky.

\textbf{K-Planes.} Aiming to further demonstrate the advantages of our method in rendering quality and speed, we compared it with K-Planes\cite{fridovich2023k}. Unlike NeRF-based methods, K-Planes uses a hybrid 3D representation that achieves lower storage and faster rendering. Quantitative results, including metrics for storage and rendering efficiency, along with qualitative results, are presented in \cref{table_s3},\cref{table_s4},\cref{fig_s4} and \cref{fig_s5}. Our method continues to achieve superior rendering quality and speed in comparison.

\textbf{3DGS.} In \cref{fig_s4} and \cref{fig_s5}, we provide visual comparisons between 3DGS and our method. It's clear that 3DGS fails to capture scene appearance variations and handle transient objects, leading to noticeable flaws and artifacts.

\subsection {K value selection}
In our method statement, K represents the number of feature maps used for sampling. We conducted experiments under the Sacre Coeur scene to determine an effective K, setting K to 1, 2, 3, 4, 8, and 16 respectively. The experimental results are presented in \cref{table_s1}. We observe that the performance on the test set does not improve when K exceeds 3 while bringing more computation cost. Thus, we reasonably set K to 3.

\subsection {More appearance tuning results}
In \cref{fig_s1}, we present additional qualitative comparison results for appearance tuning. From the figure, it's fair to conclude that our method applies the environmental factors from the reference images to the scene more reasonably, gradually increasing their influence. In the first example from top to bottom, the colors of the buildings become strange as the weight increases in the other two methods which is against practical human understanding. In the second example, our method captures the highlights on the pillars better and enhances them as the weight increases.

\begin{table}[tb]
  \caption{Quantitative results on the test set of four NeRF-OSR scenes. Our method outperforms other methods across all scenes on all metrics. 
  }
  \label{table_s4}
  \centering
  \resizebox{1.0\textwidth}{!}{
  \begin{tabular}{@{}ccccccccccccc@{}}
     \toprule
     & \multicolumn{3}{c}{stjohann} & \multicolumn{3}{c}{lwp}  &  \multicolumn{3}{c}{st}  & \multicolumn{3}{c}{europa} \\
      \cline{2-4}    \cline{5-7}     \cline{8-10}  \cline{11-13}
     & PSNR$\uparrow$ & SSIM$\uparrow$ & LPIPS$\downarrow$ & PSNR$\uparrow$ & SSIM$\uparrow$ & LPIPS$\downarrow$ & PSNR$\uparrow$ & SSIM$\uparrow$ & LPIPS$\downarrow$ & PSNR$\uparrow$ & SSIM$\uparrow$ & LPIPS$\downarrow$\\
    \midrule
   3DGS   & 17.32 & 0.7430 & 0.313 & 15.43 & 0.7000 & 0.3360 
   & 15.23 & 0.5920 & 0.4240 
   & 16.32 & 0.6520 & 0.3270\\
   K-planes-hybrid   
   & 20.39 & 0.7548 & 0.3366 
   & 21.65 & 0.7482 & 0.3397 
   & 19.66 & 0.6151 & 0.4361 
   & 18.75 & 0.6487 & 0.4549\\
   NeRF-W & 21.38 & 0.8200 & 0.1940 & 21.29 & 0.7610 & 0.2950 
   & 19.68 & 0.6310 & 0.4010 
   & 19.55 & 0.6870 & 0.3470\\
   Ha-NeRF& 19.93 & 0.7870 & 0.2100 & 21.32 & 0.7560 & 0.2780 
   & 20.56 & 0.6360 & 0.3780 
   & 18.76 & 0.6610 & 0.3480\\
   CR-NeRF& 22.27 & 0.8350 & 0.1750 & 22.61 & 0.7850 & 0.2580 
   & 21.67 & 0.6610 & 0.3600 
   & 19.92 & 0.6960 & 0.3100\\
   GS-W (Ours)   & \textbf{26.23} & \bf0.8927 & \bf0.1133 & 
   \bf24.44 & \bf0.8272 & \bf0.2082 & 
   \bf22.45 & \bf0.6900 & \bf0.3147 &
   \bf22.04 & \bf0.7577 & \bf0.2309\\
   \bottomrule
  \end{tabular}}
\end{table}

\begin{figure}[tb]
  \centering
  \includegraphics[width=\textwidth]{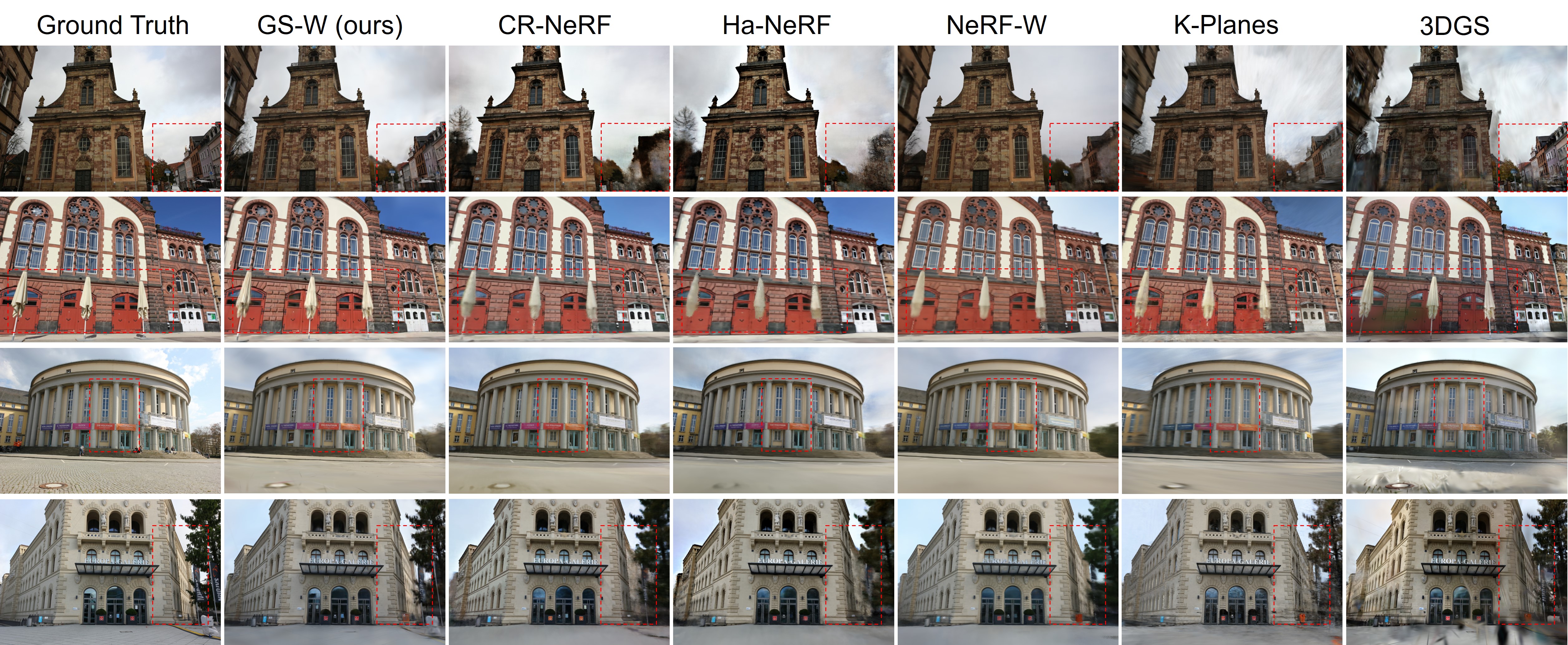}
  \caption{Qualitative comparison results on the NeRF-OSR dataset test set are shown for the stjohann, lwp, st, and europa scenes. Our method reconstructs the scenes with greater detail compared to other methods.
  }
  \label{fig_s5}
\end{figure}

\begin{figure}[tb]
  \centering
  \includegraphics[width=\textwidth]{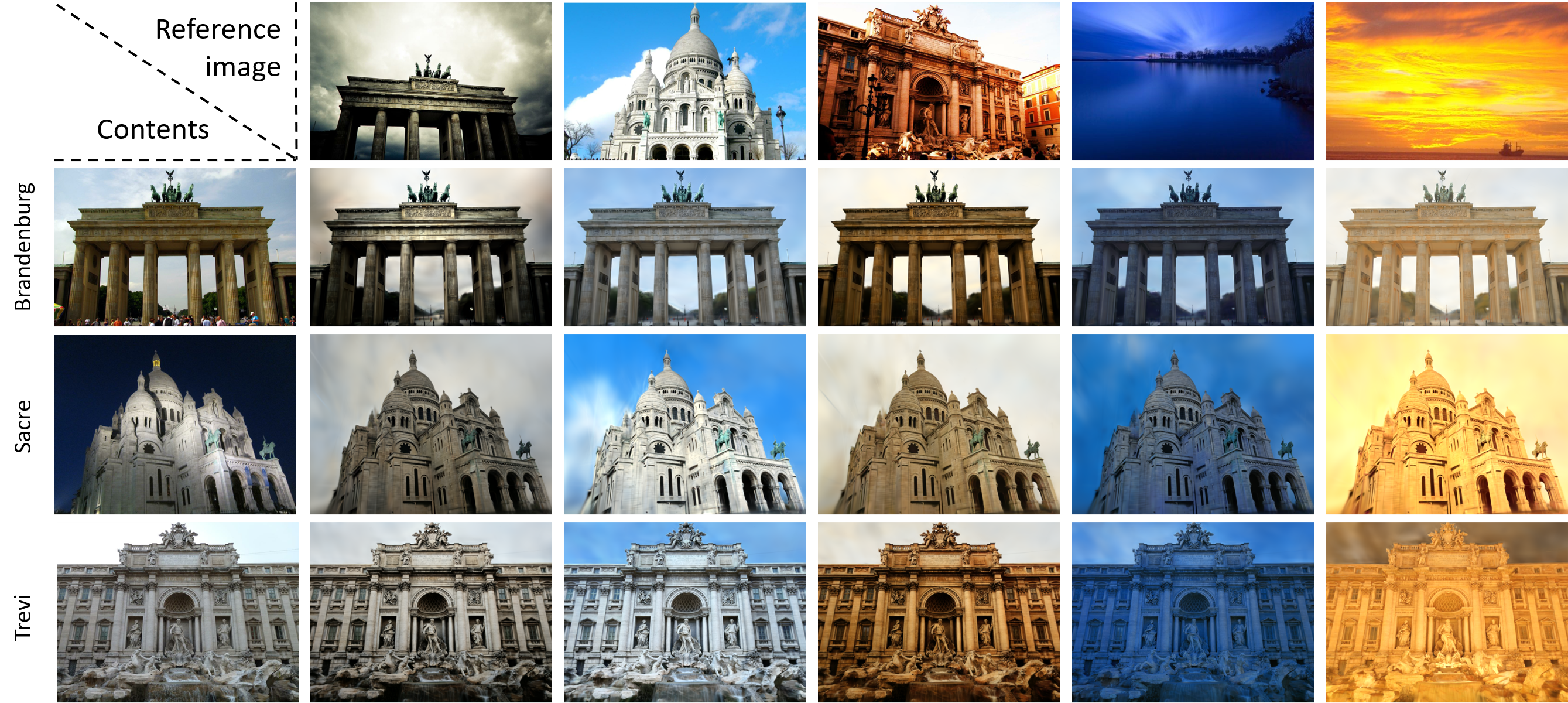}
  \caption{Style transfer from unseen images or across scenes for the three scenes. The first row represents the reference images used for extracting appearance features, while the subsequent three rows depict images rendered based on the views of the content images and the corresponding appearance features of the reference images. This demonstrates that our method can effectively perform style transfer.
  }
  \label{fig_s3}
\end{figure}

\subsection {Synthetic Lego dataset}
Following NeRF-W and Ha-NeRF, we further compared GS-W with them on the synthetic Lego dataset to validate our method's effectiveness. To ensure a fair comparison, we followed their setup by manually introducing perturbations such as colors and occluders in the training set of the synthetic Lego dataset to simulate scenarios that might be encountered in the wild, as the first column in \cref{fig_s2}. The training set comprises 100 images, while the test set consists of 200 images. During testing, only a single unperturbed training image is used as the reference image to extract features. Subsequently, novel views of the test images are rendered, and metrics are calculated against the test images.

As CR-NeRF does not provide hyperparameter settings for this dataset, the attempts to use its default parameters from other datasets result in training crashes and produce poor and peculiar results. Therefore, we only compare with NeRF-W, Ha-NeRF, and 3DGS on this dataset. Due to the limited size of the Lego training dataset and each view being influenced by a fixed color, we decode color using position instead of view direction to prevent dependency on the latter in appearance modeling. During training, without initialized point clouds, we adjust the gradient threshold for Gaussian point densification to $1.5 \times 10^{-4}$ to allow the model to generate more valid points. Other hyperparameters remain the same, with training 20k iterations.

Quantitative and qualitative results are presented in \cref{table_s2} and \cref{fig_s2}, respectively. We can observe that 3DGS performs poorly while our method performs the best under scenarios with added color and occluder perturbations. This testing process, conducted on novel viewpoints rather than on the original reference image viewpoints, also partially validates the multi-view consistency of GS-W and its ability to generalize single-image features to new viewpoints.

\subsection {NeRF-OSR dataset} We further evaluate the robustness of GS-W using the NeRF-OSR dataset\cite{rudnev2022nerf}. We experiment with four scenes - europa, lwp, st, and stjohann -  with 12.5\% of images from each sequence as the test set. Each training set contains approximately 350 images, and each test set contains about 50 images.

We compared our method with NeRF-W, Ha-NeRF, CR-NeRF, 3DGS, and K-Planes, using their default settings. Evaluation results on the test set, shown in \cref{table_s4} and \cref{fig_s5}, demonstrate that our method consistently achieves the best performance. Other methods struggle to capture local details, and 3DGS fails to handle variations in scene appearance and transient occlusions. In contrast, our method accurately captures local details and effectively handles transient occlusions.

Despite our method achieving good performance, this dataset, not sourced from the internet, features limited appearance variations per scene. Consequently, the reconstructed scenes may exhibit inconsistencies in appearance across different viewpoints. In the future, training a more generalized dynamic appearance extractor across multiple scenes could effectively address and improve this issue.

\section {Style transfer}
Similar to CR-NeRF, our method can also transfer style from images to the scene, as shown in \cref{fig_s3}. However, for style transfer across scenes or from unseen images, the absence of camera poses from the provided reference images makes it difficult to map Gaussian points to the projection feature map for sampling. Meanwhile, in style transfer, there lack of physical significance to using the projection sampling method. Therefore, when performing style transfer from unseen images, we set all features $f^P$ sampled from the projection feature map to 0. The results demonstrate that our method can effectively transfer the style provided by reference images to the scene.

\end{document}